\documentclass{article}
\usepackage{nips06,times}
\usepackage{amsmath}
\usepackage{amsfonts}
\usepackage{graphicx}

\title{Breast Cancer Diagnosis via Classification Algorithms}

\author{
Reihaneh~Entezari \\Department of Statistical Sciences\\ University of Toronto\\ 
\texttt{entezari@utstat.utoronto.ca} \\ 2013
}

\begin{document}

\maketitle

\begin{abstract}
In this paper, we analyze the Wisconsin Diagnostic Breast Cancer Data using Machine Learning classification techniques, such as the SVM, Bayesian Logistic Regression (Variational Approximation), and K-Nearest-Neighbors. We describe each model, and compare their performance through different measures. We conclude that SVM has the best performance among all other classifiers, while it competes closely with the Bayesian Logistic Regression that is ranked second best method for this dataset.
\end{abstract}

\section{Introduction}
Breast Cancer is the most common cancer among women, all around the world. There are two types of breast mass, \textbf{"Benign"} and \textbf{"Malignant"} , where benign is the non-cancerous type and malignant is the cancerous one. According to the World Health Organization, about 502,000 deaths (per year) worldwide are caused by breast cancer. The most effective way to reduce breast cancer deaths is to detect it earlier. Mammography and a Fine Needle Aspiration biopsy (FNA) are the most common diagnostic tools, which may be time consuming and cost-effective. Thus the automatic diagnosis of breast cancer becomes important, while attracting many researchers from all over the world to find an accurate and reliable diagnosis (classification) method. In Machine Learning content, there are several different methods for classification, such as SVM (Support Vector Machines) which is one of the most popular ones, Bayesian Logistic Regression (which we will use Variational Approximation here), and K-Nearest-Neighbors, etc. \\\\
SVM attempts to find the best hyperplane (by maximizing the margin) that separates the data into the two classes benign and malignant, and thus is a powerful method since it is flexible with the selection of kernel functions that map the data into higher dimension so that they are separable. Bayesian logistic regression is also another popular classification method where a prior is defined for the parameters of the model to find the posterior. After finding the posterior, we can obtain the predictive distribution and use it to predict for new test cases by defining a decision boundary.
K-Nearest-Neighbors is one of the simplest classification methods, also called as the \textit{lazy learning} where an object is assigned to the class most common amongst its \textit{k} nearest neighbours (\textit{k} being a positive integer).
\\\\
Previous work on breast cancer diagnosis using classifier techniques, have shown that usually SVM has a higher accuracy performance compared to other classifiers, although in a few, Logistic Regression was also at a high level of accuracy (depending on the data set). On the other hand, although K-NN may not be the best classifier in general,  but as we will see, it shows a desirable performance for this breast cancer diagnosis problem.  In this paper, we have collected the Wisconsin Diagnostic Breast Cancer Data (WDBC) from the UCI Machine Learning Repository to apply the methods described above. We will see that our results are in-line with previous conclusions, although a Bayesian approach to Logistic Regression will significantly increase the performance compared to the Logistic Regression.

\section{Wisconsin Diagnostic Breast Cancer Data (WDBC)}

The Wisconsin Diagnostic Breast Cancer Data consists of 569 instances with 32 attributes such as ID, diagnosis statement (B=benign, M=malignant), and 30 real-valued features that were computed from a digitized image of a fine needle aspirate (FNA) of a breast mass. Ten real-valued  features were computed for each cell nucleus: \\\\
1) radius, 2) texture, 3) perimeter, 4) area, 5) smoothness, 6) compactness, 7) concavity, \\ 8) concave points, 9) symmetry,   10) fractal dimension. \\\\
The mean, standard error, and "worst" or largest ( mean of the three largest values ) of these features were computed for each image, resulting 30 features. There are 357 benign and 212 malignant, which we will randomly split the 569 data into 80\% training set and 20\% test set to check performance.

\section{Methods}
Here we will briefly describe each classification method:

\subsection{SVM (Support Vector Machines)}
Given some data ~\begin{math}{ D = \{~ (x_i,y_i) ~ | ~x_i \in \mathbb{R}^{p}, y_i \in \{-1,1\} ~\}_{i=1}^{n} } \end{math} where \begin{math} y_i \end{math} is indicating the class of feature vector \begin{math} x_i \end{math}, the goal is to find a hyperplane that separates the two classes by maximizing the margin, which is the minimum distance from the data points to the hyperplane. In general, a hyperplane is defined as \begin{math} ~~w.x - b =0 ~~\end{math} where \begin{math} w \end{math} is the normal vector to the hyperplane and ~\begin{math} w.x\end{math}~ is the inner product of \begin{math}w\end{math} and \begin{math}x\end{math}. The parameter \begin{math} \frac{b}{||w||} \end{math} specifies the distance of the hyperplane from the origin along the normal vector \begin{math}w\end{math}. So, we have to find the best \begin{math}w\end{math} and \begin{math}b\end{math} such that the margin is maximized. Hence if the training data are linearly separable, the optimizing problem will be: \begin{center}
\begin{equation} Minimize_{~w,b}~~~\frac{1}{2} ||w||^2 
~~~~~subject~ ~to~ ~~~y_i(w.x_i - b) \geq 1
\end{equation} 
\end{center}
Where the solution to \begin{math}w\end{math} will be \begin{math} w^* = \sum_{i}{\alpha_iy_ix_i} \end{math}.\\\\
There is a modified maximum margin optimization that allows misclassification of training points, when there exists no hyperplane that can completely separate the training set. It is called the \textit{Soft Margin} method, which will choose a hyperplane that separates the data as cleanly as possible, while maximizing the distance to the nearest cleanly separated points. In this method, new parameters, \begin{math} \xi_i\end{math}, are added to measure the degree of misclassification for each data \begin{math}x_i\end{math}, by :
\begin{center}
 \begin{math} y_i(w.x_i - b) \geq 1- \xi_i~~~~~~1\leq i \leq n \end{math}
\end{center}
The new optimization will become a trade off between a large margin and a small error penalty. If the penalty function is linear, the optimization problem changes to:
\begin{center}
\begin{equation}
Minimize_{~w,\xi,b}~~~\{ \frac{1}{2}||w||^2 + C \sum_{i=1}^{n}\xi_i \}
~~~~subject ~~to~~~~~ y_i(w.x_i - b) \geq 1- \xi_i,~~~\xi_i\geq 0
\end{equation}
\end{center}
Where the constant \textit{C} controls the trade off between complexity of decision rule and frequency of error. By using Lagrange Multipliers \begin{math}\alpha_i\end{math}, the dual form of the previous optimization problem will be:
\begin{center}
\begin{equation}
Maximize_{~\alpha_i}~\{ \sum_{i=1}^{n}{\alpha_i} - \frac{1}{2}\sum_{i,j}{\alpha_i\alpha_jy_iy_jk(x_i,x_j)} \}
~~~~~~subject ~~to~~~~~0\leq \alpha_i \leq C,~~~and~~~\sum_{i=1}^{n}{\alpha_iy_i}=0
\end{equation}
\end{center}
Where \begin{math} k(x_i,x_j) = x_i.x_j \end{math} is called the kernel function (here it's linear). In general, kernel functions are defined as \begin{math} k(x_i,x_j) = \phi(x_i).\phi(x_j) \end{math}, where \begin{math}\phi \end{math} is a function that maps the data to a higher dimensional space called the \textit{feature space}, to make a better look of the data. Function \begin{math}\phi \end{math} doesn't have to be known since just their inner product is important in the optimization problem. Below are some possible kernel functions:
\begin{center}
\begin{itemize}
\item Polynomial: \begin{math} k(x_i,x_j) = (x_i.x_j)^d \end{math}
\item Gaussian radial basis function: \begin{math} k(x_i,x_j) = e^{-\gamma||x_i-x_j||^2} \end{math} where \begin{math}\gamma > 0 \end{math}, \\ we can also assume \begin{math} \gamma = \frac{1}{2\sigma^2} \end{math}
\item Hyperbolic tangent: \begin{math} k(x_i,x_j) = tanh(\kappa x_i.x_j + c)\end{math} for some (not every) \begin{math}\kappa > 0\end{math} and \begin{math} c < 0\end{math}
\end{itemize}
\end{center}
After finding the optimum \begin{math}w^*\end{math} and \begin{math}b^*\end{math}, we can classify a new test case \begin{math}x^*\end{math} by \begin{math} sign(w^*.\phi(x) + b^*)\end{math} which is equivalent to \begin{math} sign(\sum_{i=1}^{n}{\alpha_iy_ik(x_i,x^*)} + b^*)\end{math}. 

\subsection{Bayesian Logistic Regression}
The model known as \textit{Logistic Regression}, which is a model for classification, is as below:
\begin{center}
\begin{equation}
p(C_1| \phi) = \sigma (w^{T} \phi)
\end{equation}
\end{center}
with \begin{math} p(C_2|\phi) = 1 - p(C_1|\phi)\end{math}, where \begin{math} \sigma(x) = \frac{1}{1+e^{-x}} \end{math} is the logistic sigmoid function and  \begin{math}\phi \end{math} is the feature space. Then, the likelihood function for the data \begin{math} \{ \phi_i, t_i \}_{i=1}^{n}\end{math}, where \begin{math}t_i \in \{0,1\} \end{math} and \begin{math}\phi_i = \phi(x_i) \end{math}, will be:
\begin{center}
\begin{equation}
p(t|w) = \prod_{i=1}^{n}{p(C_1|\phi_i)^{t_i} {(1-p(C_1|\phi_i))}^{1-t_i}}
\end{equation}
\end{center}
Where \begin{math} t = (t_1,...,t_n)^T \end{math}. To predict for new test cases, we need to estimate the parameter \begin{math} w \end{math}, which is a \begin{math}p \end{math} dimensional vector, according to \begin{math} p\end{math} features for each training data. One way is to estimate it directly, using any iterative optimization scheme, such as Newton-Raphson, Gradient Descent, etc. Another way is to look at the bayesian approach where a prior for \begin{math} w \end{math} is chosen to obtain the posterior. Usually Gaussian priors are considered, and since~~\begin{math} posterior \propto prior \times likelihood \end{math} , computing the posterior would be intractable (logistic sigmoid and a Gaussian). On the other hand, the posterior can be approximated using various methods, such as Laplace approximation, MCMC, etc. Here we will approximate the posterior using the variational approximation. We will assume the prior is Gaussian, \begin{math}p(w) = \mathcal{N}(w| m_0,S_0) \end{math}. \\\\
In the variational framework, we wish to maximize a lower bound on the marginal likelihood:
\begin{center}
\begin{equation}
p(t) = \int { p(t|w)p(w)dw} = \int{ [~ \prod_{i=1}^{n} p(t_i|w)~] ~p(w) dw}
\end{equation}
\end{center}
Before we proceed to the next step, note that we also have the following property for each  \begin{math}i\end{math}:
\begin{center}
\begin{equation}
p(t_i|w) = \sigma(a_i)^{t_i}\{1-\sigma(a_i)\}^{1-t_i} = (\frac{1}{1+e^{-a_i}})^{t_i}(1-\frac{1}{1+e^{-a_i}})^{1-t_i} =  e^{a_it_i}\frac{e^{-a_i}}{1+e^{-a_i}} = e^{a_it_i}\sigma(-a_i)
\end{equation}
\end{center}
Where \begin{math}a_i=w^T\phi_i\end{math}. Now to find a lower bound on \begin{math}p(t)\end{math}, from (6) and (7), we need to first use the variational lower bound for the logistic sigmoid function [1], hence:
\begin{center}
\begin{equation}
p(t_i|w) = e^{a_it_i}\sigma(-a_i) ~\geq~  e^{a_it_i} \sigma(\xi_i) e^{\{ -\frac{(a_i + \xi_i)}{2} - \lambda(\xi_i)({a_i}^{2} - {\xi_i}^2)\}}
\end{equation}
\end{center}
where \begin{math}\lambda(\xi) = \frac{1}{2\xi}[\sigma(\xi)-\frac{1}{2}]\end{math} and \begin{math}\xi_i\end{math} are the variational parameters for each data point. Now by (6) and (8), we have :
\begin{center}
\begin{equation}
p(t|w)p(w) \geq h(w,\xi)p(w)
\end{equation}
\end{center}
where \begin{math} \xi = \{ \xi_1,..., \xi_n\} \end{math} and \begin{math} h(w,\xi) = \prod_{i=1}^{n}{\sigma(\xi_i)e^{\{w^T\phi_it_i - \frac{(w^T\phi_i + \xi_i)}{2} - \lambda(\xi_i)([w^T\phi_i]^2 - {\xi_i}^2)\}}}\end{math}\\\\
Now, by taking the logarithm of both sides of (9), we will see that the right hand side will be a quadratic function of \begin{math}w\end{math}. Hence the variational approximation to the posterior distribution of \begin{math}w\end{math} will have a Gaussian form as below:
\begin{center}
\begin{equation}
q(w) = \mathcal{N}(w|\mu_N,S_N)
\end{equation}
\end{center}
\begin{center}
\begin{equation}
S_N = (S_0^{-1} + 2\sum_{i=1}^{n}{\lambda(\xi_i)\phi_i{\phi_i}^T})^{-1}~~~~\&~~~~~\mu_N = S_N(S_0^{-1}m_0 + \sum_{i=1}^{n}{(t_i - \frac{1}{2})\phi_i})
\end{equation}
\end{center}
On the other hand, we need to find the variational parameters \begin{math}\xi_i\end{math} by maximizing the lower bound on the marginal likelihood given in (6). One way to approach this, is to use the EM algorithm, followed by:
\begin{center}
\begin{itemize}
\item Initialize all  \begin{math}\xi_i\end{math} s
\item E-Step: use these parameters to find the posterior distribution over w, as given in (10), (11)
\item M-Step: maximize the expected complete-data log likelihood given by \begin{math} \mathbb{Q}(\xi,\xi^{old}) = \mathbb{E}[ln~ h(w,\xi)p(w)] \end{math}. Which is equivalent to solving: \begin{math}(\xi_i^{new})^2 = \phi_i^T(S_N + \mu_N\mu_N^T)\phi_i\end{math}
\end{itemize}
\end{center}
Up to now, we have assumed that hyperparameters (in prior) are known. Hyperparameters can also be unknown, so that they can be learned from the data. Hence, we will need additional priors for the hyperparameters. Assume the prior on \begin{math}w \end{math} depends on another unknown parameter \begin{math}\alpha \end{math}. Thus, we can consider the prior for  \begin{math} w\end{math} to be Gaussian again,  \begin{math} p(w|\alpha) = \mathcal{N}(w|0, \alpha^{-1}I )\end{math}, and the prior for  \begin{math}\alpha \end{math} to be gamma,  \begin{math}p(\alpha) = Gam(\alpha|a_0,b_0) \end{math}. Now the marginal likelihood will be:
\begin{center}
\begin{equation}
p(t) = \int\int{p(t|w)p(w|\alpha)p(\alpha)dw d\alpha}
\end{equation}
\end{center}
Since this is analytically intractable, we will consider the variational distribution \begin{math} q(w,\alpha)\end{math} and the decomposition :
\begin{center}
\begin{equation}
ln~p(t) = \mathcal{L}(q) + KL(q||p)
\end{equation}
\end{center}
Where 
\begin{center}
\begin{equation}
\mathcal{L}(q) = \int\int{q(w,\alpha)~ ln~\{ \frac{p(t|w)p(w|\alpha)p(\alpha)}{q(w,\alpha)} \}dwd\alpha}
\end{equation}
\end{center}
And
\begin{center}
\begin{equation}
KL(q||p) =  - \int\int{q(w,\alpha)~ ln~\{ \frac{p(w,\alpha|t)}{q(w,\alpha)} \}dwd\alpha}
\end{equation}
\end{center}
Since \begin{math}\mathcal{L}(q) \end{math} is still intractable, we will use the local variational bound on the logistic sigmoid function (just as before). Hence we will have:
\begin{center}
\begin{equation}
ln~p(t) \geq \mathcal{L}(q) \geq \tilde{\mathcal{L}}(q,\xi) = \int \int {q(w,\alpha)~ ln~\{ \frac{h(w,\xi)p(w|\alpha)p(\alpha)}{q(w,\alpha)} \}dwd\alpha}
\end{equation}
\end{center}
Next, we assume that the variational distribution factorizes between parameters and hyperparameters: \begin{math}q(w,\alpha) = q(w)q(\alpha) \end{math}~. Now we will find each factor as below [2]:
\begin{center}
\begin{equation}
ln~q(w) = \mathbb{E}_{\alpha}{[ln~\{h(w,\xi)p(w|\alpha)p(\alpha)\}]} + constant
\end{equation}
\end{center}
\begin{center}
\begin{equation}
ln~q(\alpha) = \mathbb{E}_{w}{[ln~p(w|\alpha)]} + ln~p(\alpha) + constant
\end{equation}
\end{center}
Omitting the algebra, \begin{math}w\end{math} will follow \begin{math}q(w) = \mathbb{N}(w|\mu_N,S_N)\end{math} and \begin{math}\alpha \end{math} will follow \begin{math}q(\alpha) = Gam(\alpha|a_N,b_N)\end{math} , with updated equations :
\begin{center}
\begin{equation}
S_N = (\mathbb{E}[\alpha]I + 2\sum_{i=1}^{n}{\lambda(\xi_i)\phi_i{\phi_i}^T})^{-1} ~~~~~~~~\&~~~~~~~\mu_N = S_N( \sum_{i=1}^{n}{(t_i - \frac{1}{2})\phi_i})
\end{equation}
\end{center}
Also
\begin{center}
\begin{equation}
a_N = a_0 + \frac{P}{2}   ~~~~~~~~\&~~~~~~~  b_N= b_0 + \frac{1}{2}\mathbb{E}_w[w^{T}w]
\end{equation}
\end{center}

where \begin{math}P\end{math} is the dimension of feature space and  \begin{math}\mathbb{E}[\alpha]=\frac{a_N}{b_N}~,~\mathbb{E}_w[w^{T}w] = Tr(S_N + \mu \mu^T) \end{math}. We should note that the updated equations for  \begin{math}\xi_i\end{math} is the same as before.\\\\
On the other hand, the variational lower bound would be:
\begin{center}
\begin{equation}
\tilde{\mathcal{L}}(q,\xi) = \mathbb{E}_w[ln~h(w,\xi)] + \mathbb{E}_{w,\alpha}[ln~p(w|\alpha)] + \mathbb{E}_{\alpha}[ln~p(\alpha)] - \mathbb{E}_{w}[ln~q(w)]- \mathbb{E}_{\alpha}[ln~q(\alpha)] 
\end{equation}
\end{center}
With:
\begin{center}
\begin{equation}
\mathbb{E}_w[ln~h(w,\xi)]  =  \sum_{i=1}^{n}{\{ ~ln~\sigma(\xi_i) + \mu_N^{T}\phi_it_i - \frac{(\mu_N^{T}\phi_i + \xi_i)}{2} - \lambda(\xi_i)[\mu_N^{T}\phi_i\phi_i^{T}\mu_N + Tr(\phi_i\phi_i^{T}S_N) - \xi_i^{2}] ~\} } 
\end{equation}
\end{center}
\begin{center}
\begin{equation}
\mathbb{E}_{w,\alpha}[ln~p(w|\alpha)]  = -\frac{P}{2} ln(2\pi) + \frac{P}{2}(\psi(a_N) - ln~b_N) - \frac{a_N}{2b_N}(\mu_N^{T}\mu_N + Tr(S_N))
\end{equation}
\end{center}
\begin{center}
\begin{equation}
\mathbb{E}_{\alpha}[ln~p(\alpha)]  = a_0 ln b_0 + (a_0 - 1)(\psi(a_N) - ln b_N) - b_0 \frac{a_N}{b_N} - ln \Gamma(a_0)
\end{equation}
\end{center}
\begin{center}
\begin{equation}
- \mathbb{E}_{w}[ln~q(w)] = \frac{1}{2}ln |S_N| + \frac{P}{2} ln(2\pi)~~~~~\&~~~~- \mathbb{E}_{\alpha}[ln~q(\alpha)]  = ln\Gamma(a_N) - (a_N - 1)\psi(a_N) - ln b_N + a_N
\end{equation}
\end{center}
\subsubsection{Predictive distribution}
Given a new test case (in its feature space) \begin{math}\phi(x^*)\end{math}, we can compute the predictive distribution for class \begin{math}\mathcal{C}_1\end{math} by marginalizing with respect to the posterior distribution for \begin{math}p(w|t) \end{math} which is approximated with the Gaussian \begin{math}q(w) \end{math}:
\begin{center}
\begin{equation}
p(\mathcal{C}_1|\phi(x^*),t) = \int {p(\mathcal{C}_1|\phi(x^*),w) p(w|t) dw} \simeq \int {\sigma(w^{T}\phi(x^*))q(w)dw}
\end{equation}
\end{center}
with  \begin{math}p(\mathcal{C}_2|\phi(x^*),t) = 1 - p(\mathcal{C}_1|\phi(x^*),t)\end{math}. We have 
computed the above integral using the Monte Carlo technique, by sampling from the Gaussian distribution \begin{math}q(w) \end{math}.

\subsection{K-Nearest-Neighbors}
As mentioned before, in this method a test point is classified by assigning the label which is most frequent among the \begin{math}k\end{math} (a user-defined constant) training points nearest to that test point. Euclidean distance is one of the options to use as the measure of closeness (distance metric), which we will apply here. 

\section{Results}
\subsection{Error Rates}
After randomly dividing the data into 80\% training set and 20\% test set (455 Training data, 114 Test data), we have calculated the training error rate and the test error rate for each method. We should note that for SVM, the error rates for each kernel type was resulted from the best parameters chosen from a finite set of different values, by using 10-fold cross-validation. Table 1 below are showing these error rates and the number of misclassified points.

\begin{table}[h]
\caption{Testing and training error rates}
\label{sample-table}
\begin{center}
\begin{tabular}{lll}
\bf{ Methods}  & \bf{Testing Error Rates} & \bf{Training Error Rates}
\\ \hline \\
SVM (Gaussian Radial Basis)        &    0.01754386 ~ (2 miss)  &  0.01318681 ~(6 miss)     \\
SVM (Polynomial)        &    0.02631579 ~ (3 miss)  &    0.01758242 ~(8 miss)  \\
SVM (Hyperbolic Tangent)        &    0.07894737 ~(9 miss)   &   0.08791209 ~ (40 miss)  \\
K-NN (k=1)        &    0.1052632 ~(12 miss)   &    0.07252747~ (33 miss)  \\
K-NN (k=3)        &    0.0877193  ~(10 miss)  &    0.07252747 ~(33 miss)  \\
K-NN (k=10)        &    0.0877193  ~(10 miss)  &    0.06153846  ~(28 miss) \\
Logistic Regression        &    0.09649123 ~(11 miss)   &   0.07692308  ~(35 miss)  \\
Bayesian Logistic Regression (Variational Appr.)        &    0.06140351  ~(7 miss)  &   0.03296703 ~(15 miss)   \\
\end{tabular}
\end{center}
\end{table}
\begin{figure}[h]
  \centering
   \includegraphics[width=0.6\columnwidth]{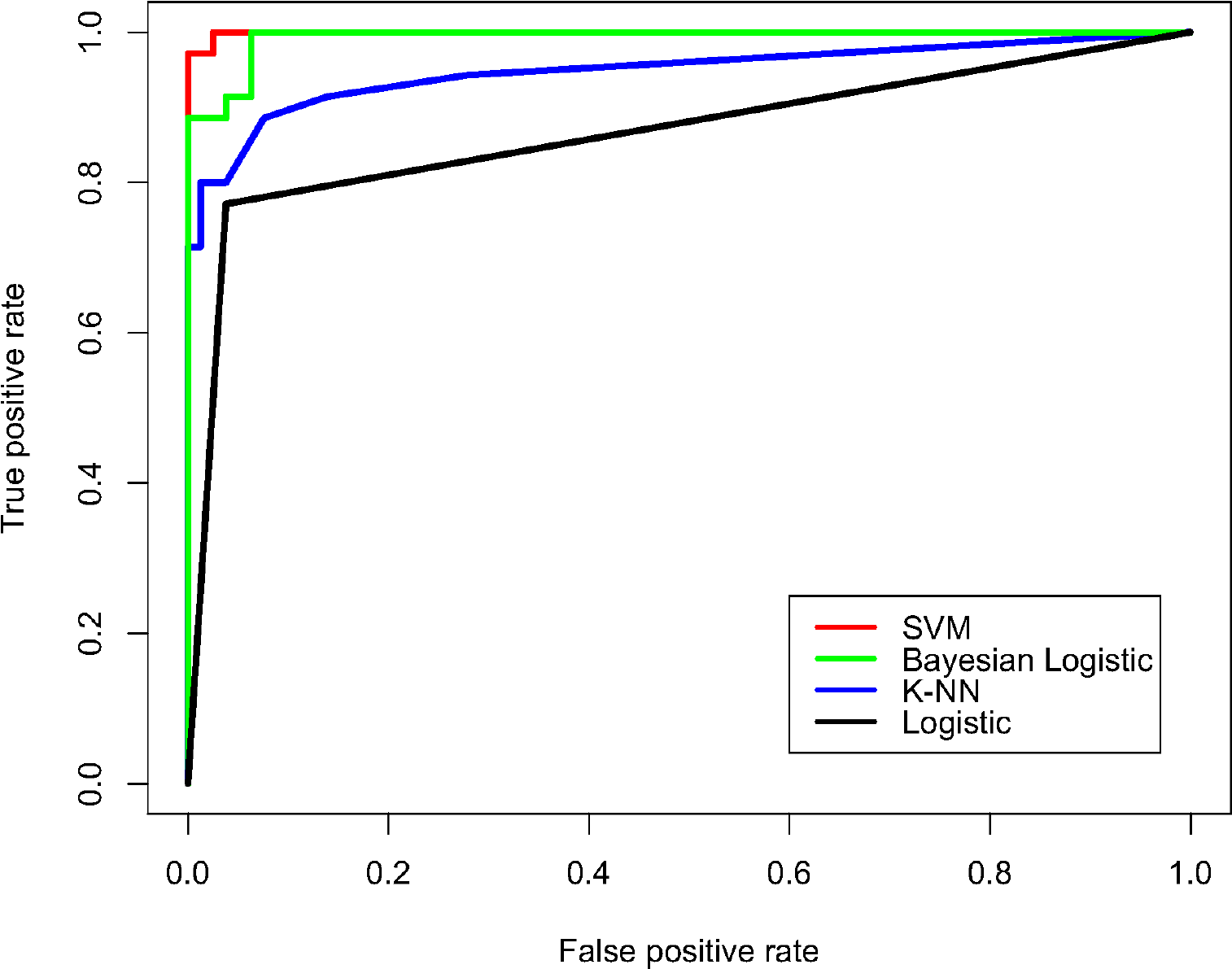}
   \caption{ROC curves}
 \end{figure}

As it is seen from the test error rates and training error rates, SVM with a Gaussian Radial Basis kernel has the smallest rate (both test and training) among the other SVMs, and K-NN with k=10 has also the smallest error rate among the other K-NNs. So to compare the performance of these methods (after selecting the best from each category, explained earlier), we see that their error rates rank from smallest to largest by: 1) SVM(Gaussian)~2) Bayesian Logistic Regression~3) K-NN(k=10)~4) Logistic Regression. To have a more accurate result, in the next section, we will look at the ROC curves and AUC to compare these 4 methods ( after selection ).
\subsection{ROC curves and AUC}
In classification, specifically diagnosis of cancer, it's important that what class the test cases are assigned to. There are four possible cases: 1) True Positive[TP] ( If new data is assigned to class positive, and it actually belongs to that class) 2) False Positive[FP] ( If it is assigned to positive class, but is actually in the negative class, in our case a person is diagnosed as cancerous, while being healthy) 3) True Negative[TN] ( when it is assigned to class negative and it actually belongs to that class) 4) False Negative[FN] ( It is assigned to class negative while it belongs to the positive class, in our case the person is predicted to be healthy, but actually has cancer, which is completely dangerous).  In the previous section, we just considered all the false positive and false negative together, and computed the error. 
To conclude more accurate results, we have to consider all these cases, where in specific, ROC curves are used for showing the "True positive Positive rate(TPR)" vs "False Positive rate(FPR)" [\begin{math} TPR = \frac{TP}{TP + FN}\end{math} and \begin{math} FPR = \frac{FP}{FP + TN}\end{math}]. ROC curves represent relative tradeoffs between benefits (TP) and costs (FP).
As it is seen in Figure 1, the distance of the ROC curves of SVM, Bayesian Logistic Regression, K-NN and Logistic Regression from the diagonal line are ordered from largest to smallest, respectively, showing performance (from highest to least). So again the same order of performance is resulted here with the ROC curves. Another way to measure the performance is to look at the AUC values (which are the area under the ROC curves) and also the Accuracy measurement. They are given in Table 2.

 \begin{table}[h]
\caption{AUC (area under ROC curves)}
\label{sample-table}
\begin{center}
\begin{tabular}{lll}
\bf{ Methods}  & \bf{AUC} & \bf{Accuracy}
\\ \hline \\
SVM        &   0.9992767   &   98\%    \\
Bayesian Logistic Regression        &    0.9934901  &   94\%   \\
K-NN        &    0.9484629  &    91\% \\
Logistic Regression        &    0.8667269   &   90\%  \\
\end{tabular}
\end{center}
\end{table}

Again as it is seen from Table 2, the AUCs and accuracies are ordered from largest to smallest with the same orders of methods. The accuracy vs cut-off graphs for each method are also shown in Figure 2. Since our decision boundary was probability 0.5 ( or the sign function in SVM), we see that all methods (except Logistic Regression) have their highest accuracy (Figure 2).

 \begin{figure}[h]
  \centering
   \includegraphics[width=0.7\columnwidth]{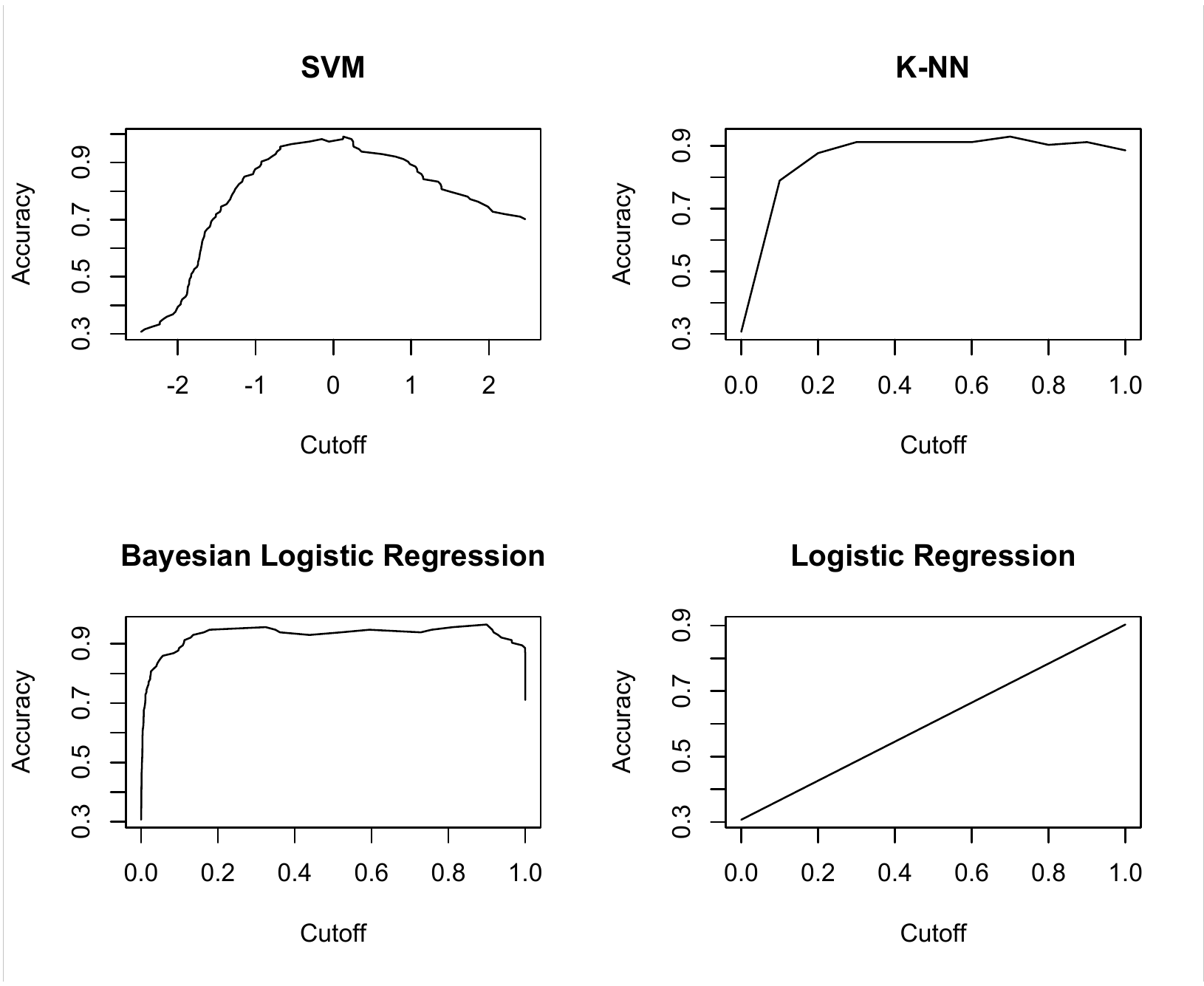}
   \caption{Accuracy vs Cutoff}
 \end{figure}

\section{Conclusion} 
Finally, to summarize the results, we started by looking at test error rates, ROC curves and at last AUC and accuracy values for each method. From the comparison, we see that SVM has the least error rate (highest Accuracy), as well as the highest AUC value, so it has the highest performance among the others (for this data set). On the other hand, Bayesian Logistic Regression is ranked second best, which has improved the performance of usual Logistic Regression. And at last, K-NN and Logistic Regression were ranked third and fourth respectively. But in general, they were all good classifiers since they all had high accuracy rates. The study of feature selection can be considered as a future direction for this dataset.

\subsubsection*{References}

\small{
[1] Christopher M. Bishop, {\it Pattern Recognition and Machine Learning}, pp. 497-505 , pp. 466. Cambridge.

[2] Katie Planey, {\it Machine Learning Approaches to Breast Cancer Diagnosis and Treatment Response Prediction}. Stanford Biomedical Informatics.

[3] Gouda I. Salama, M.B.Abdelhalim, and Magdy Abd-elghany Zeid (2012), {\it Breast Cancer Diagnosis on Three Different Data Sets Using Multi-Classfiers}. Computer Science, Arab Academy for Science Technology \& Maritime Transport, Cairo, Egypt.

[4] Y.Ireaneus Anna Rejani, Dr.S.Thamarai Selvi [2009], {\it Early Detection Of Breast Cancer Using SVM Classifier Technique}. Noorul Islam College of Engineering, Tamilnadu, India.

[5] Lukasz Jelen , Thomas Fevens, Adam Krzyz  Ak [2008], {\it Classification Of Breast Cancer Malignancy Using Cytological Images Of Fine Needle Aspiration Biopsies} . Department of Computer Science and Software Engineering Concordia University, Montreal, Quebec, Canada.

[6] Manish Sarkar and Tze-Yun Leong, \it{Application of K-Nearest-Neighbors Algorithm on Breast Cancer Diagnosis problem}. Department of Computer Science, School of Computing, The National university of Singapore.

[7] Tommi S. Jaakkola and Michael I. Jordan [1999], {\it Bayesian Parameter Estimation Via Variational Methods}. Dept. of Elec. Eng. \& Computer Science, Massachusetts Institute of Technology, Cambridge, MA, USA. 

[8] Tom Fawcett [2006], {\it An Introduction To ROC Analysis}. Institute for the Study of Learning and Expertise, 2164 Staunton Court, Palo Alto, CA 94306, USA.
}

\end{document}